\documentclass[11pt,a4paper]{article}
\usepackage[hyperref]{emnlp2018}
\usepackage{times}
\usepackage{latexsym}
\usepackage{amsfonts}
\usepackage{amsmath}
\usepackage{multirow}

\usepackage{url}

\aclfinalcopy 

\title{A Dataset for Document Grounded Conversations}

\author{Kangyan Zhou, Shrimai Prabhumoye, Alan W Black\\
  Carnegie Mellon University, Pittsburgh, PA, USA\\
  {\tt \{kangyanz, sprabhum, awb\}@cs.cmu.edu}
  }

\date{}

\begin{document}
\maketitle
\begin{abstract}
This paper introduces a document grounded dataset for conversations.
We define ``Document Grounded Conversations'' as conversations that are about the contents of a specified document. In this dataset the specified documents were 
Wikipedia articles about popular movies.
The dataset contains 4112 conversations with an average of 21.43 turns per conversation.
This positions this dataset to not only provide a relevant chat history while generating responses but also provide a source of information that the models could use.
We describe two neural architectures that provide benchmark performance on the task of generating the next response.
We also evaluate our models for engagement and fluency, and find that the information from the document helps in generating more engaging and fluent responses.

\end{abstract}

\section{Introduction}

At present, dialog systems are considered to be either task-oriented, where a specific task is the goal of the conversation (e.g. getting bus information or weather for a particular location);
or non-task oriented where conversations are more for the sake of themselves, be it entertainment or passing the time.
Ultimately, we want our agents to smoothly interleave between task-related information flow and casual chat for the given situation.
There is a dire need of a dataset which caters to both these objectives.

\citet{serban2015survey} provide a comprehensive list of available datasets for building end-to-end conversational agents. 
Datasets based on movie scripts \cite{lison2016opensubtitles2016,danescu2011chameleons} contain artificial conversations.
The Ubuntu Dialogue Corpus \cite{lowe2015ubuntu} is based on technical support logs from the Ubuntu forum.
The Frames dataset \cite{asri2017frames} was collected to solve the problem of frame tracking.
These datasets do not provide grounding of the information presented in the conversations.
\citet{zhang2018personalizing} focuses on personas in dialogues: each worker has a set of predefined facts about the persona that they can talk about. 
Most of these datasets lack conversations with large number of on-topic turns.

We introduce a new dataset which addresses the concerns of grounding in conversation responses, context and coherence in responses.
We present a dataset which has real human conversations with grounding in a document.
Although our examples use Wikipedia articles about movies, we see the same techniques being valid for other external documents such as manuals, instruction booklets, and other informational documents.
We build a generative model with and without the document information and find that the responses generated by the model with the document information is more engaging (+7.5\% preference) and more fluent (+0.96 MOS). 
The perplexity also shows a 11.69 point improvement.

\section{The Document Grounded Dataset}
To create a dataset for document grounded conversations, we seek the following things:
(1) A set of documents
(2) Two humans chatting about the content of the document for more than 12 turns. 
We collected conversations about the documents through Amazon Mechanical Turk (AMT). 
We restrict the topic of the documents to be movie-related articles to facilitate the conversations.  
We initially experimented with different potential domains. Since movies are engaging and widely known, people actually stay on task when discussing them. 
In fact in order to make the task interesting, we offered a choice of movies to the participants so that they are invested in the task.

\subsection{Document Set Creation}

We choose Wikipedia (Wiki) \footnote{https://en.wikipedia.org} 
articles to create a set of documents 
$\boldsymbol{D} = \{d_1, \ldots, d_{30}\}$ for grounding of conversations. 
We randomly select 30 movies, covering various genres like 
thriller, super-hero, animation, romantic, biopic etc.
We extract the key information provided in the Wiki article and divide it into four separate sections.
This was done to reduce the load of the users to read, absorb and discuss the information in the document.
Hence, each movie document $d_i$ consists of four sections $\{s_1, s_2, s_3, s_4\}$ corresponding to basic information and three key scenes of the movie.
The basic information section $s_1$ contains data from the Wikipedia article in a standard form such as year, genre, director. 
It also includes a short introduction about the movie, ratings from major review websites, and some critical responses. 
Each of the key scene sections $\{s_2, s_3, s_4\}$ contains one short paragraph from the plot of the movie. 
Each paragraph contains on an average 7 sentences and 143 words.
These paragraphs were extracted automatically from the original articles, and were then lightly edited by hand to make them of consistent size and detail.
An example of the document is attached in Appendix.

\subsection{Dataset Creation}

To create a dataset of conversations which uses the information from the document, involves the participation of two workers.
Hence, we explore two scenarios: (1) Only one worker has access to the document and the other worker does not and (2) Both the workers have access to the document.
In both settings, they are given the common instructions of chatting for at least 12 turns.

\paragraph{Scenario 1: One worker has document.} 
In this scenario, only one worker has access to the document. 
The other worker cannot see the document. 
The instruction to the worker with the document is: \textit{Tell the other user what the movie is, and try to persuade the other user to watch/not to watch the movie using the information in the document}; 
and the instruction to the worker without the document is: \textit{After you are told the name of the movie, pretend you are interested in watching the movie, and try to gather all the information you need to make a decision whether to watch the movie in the end}. 
An example of part of the dialogue for this scenario is shown in Table \ref{table:scenario_1}.

\begin{table}[t!]
\centering
\begin{tabular}{| p{0.7cm} p{6.3cm} |}
\hline
user2: & Hey have you seen the inception? \\
user1: & No, I have not but have heard of it. \\
& What is it about \\
user2: & It's about extractors that perform \\
& experiments using military technology \\
& on people to retrieve info about their \\ 
& targets. \\
\hline
\end{tabular}
\caption{An example conversation for scenario 1. User 1 does not have access to the document, while User 2 does. The full dialogue is attached in the Appendix.}
\label{table:scenario_1}
\end{table}

\paragraph{Scenario 2: Both workers have document.} 
In this scenario, both the workers have access to the same Wiki document. 
The instruction given to the workers are: \textit{Discuss the content in the document with the other user, and show whether you like/dislike the movie}. 
An example of the dialogue for this scenario is shown in Table \ref{table:scenario_2}.

\begin{table}[t!]
\centering
\begin{tabular}{| l l |}
\hline
User 2: & I thought The Shape of Water was \\
& one of Del Toro's best works. \\ 
& What about you?\\
User 1: & Did you like the movie?\\
User 1: & Yes, his style really extended the story.\\
User 2: & I agree. He has a way with fantasy \\
& elements that really helped this story \\
& be truly beautiful.\\
\hline
\end{tabular}
\caption{An example conversation for scenario 2. Both User 1 and User 2 have access to the Wiki document. The full dialogue is attached in the Appendix.}
\label{table:scenario_2}
\end{table}

\paragraph{Workflow:} 
When the two workers enter the chat-room, they are initially given only the first section on basic information $s_1$ of the document $d_i$.
After the two workers complete 3 turns (for the first section 6 turns is needed due to initial greetings), 
the users will be shown the next section.
The users are encouraged to discuss information in the new section, but are not constrained to do so. 



\begin{table*}[h!]
\centering
\begin{tabular}{|l|l|l|}
\hline
Dataset                              & \# Utterances & Avg. \# of Turns \\ \hline
CMU\_DoG                                 & 130，000       & 31              \\ \hline
Persona-chat \cite{zhang2018personalizing}   & 164,356       & 14               \\ \hline
Cornell Movie \cite{Danescu-Niculescu-Mizil+Lee:11a} & 304,713       & 1.38             \\ \hline
Frames dataset \cite{asri2017frames}        & 19,986        & 15               \\ \hline
\end{tabular}
\caption{Comparison with other datasets. The average number of turns are calculated as the number  of utterances divided by the number of conversations for each of the datasets.}
\label{table:stat_comparison}
\end{table*}

\begin{table*}[t!]
\centering
\begin{tabular}{|l|l|l|l|l|}
\hline
	     & Rating 1	 &  Rating 2 &   Rating 3 & Rating 2\& 3 \\ \hline
Total \# of conversations & 1443 & 2142 &   527 & 2669 \\ \hline
Total \# of utterances & 28536 & 80104 &   21360 & 101464 \\ \hline
Average \# utterances/conversation & 19.77(13.68) &	35.39(8.48) &	40.53(12.92) & 38.01(9.607) \\ \hline
Average length of utterance & 7.51(50.19) &	10.56(8.51) &	16.57(15.23) & 11.83(10.58) \\ \hline
\end{tabular}
\caption{The statistics of the dataset. Standard deviation in parenthesis.}
\label{table:dataset_statistics}
\end{table*}

\subsection{Dataset Statistics}

The dataset consists of total 4112 conversations with an average of 21.43 turns.
The number of conversations for scenario 1 is 2128 and for scenario 2 it is 1984.
We consider a turn to be an exchange between two workers (say $w1$ and $w_2$). 
Hence an exchange of $w_1, w_2, w_1$ has 2 turns ($w_1$, $w_2$) and ($w_2$, $w_1$).
We show the comparison of our dataset as \textbf{CMU} \textbf{Do}cument \textbf{G}rounded Conversations (CMU\_DoG) with other datasets in Table \ref{table:stat_comparison}.
One of the salient features of CMU\-DoG dataset is that it has mapping of the conversation turns to each section of the document, which can then be used to model conversation responses.
Another useful aspect is that we report the quality of the conversations in terms of how much the conversation adheres to the information in the document.

\paragraph{Split Criteria:} 
We automatically measure the quality of the conversations using BLEU \cite{papineni2002bleu} score.
We use BLEU because we want to measure the overlap of the turns of the conversation with the sections of the document.
Hence, a good quality conversation should use more information from the document than a low quality conversation.
We divide our dataset into three ratings based on this measure.
The BLEU score is calculated between all the utterances $\{x_1, \ldots, x_n\}$ of a conversation $C_i$ and the document $d_i$ corresponding to $C_i$. 
We eliminate incomplete conversations that have less than 10 turns.
The percentiles for the remaining conversations are shown in Table \ref{table:bleu_distribution}. 
We split the dataset into three ratings based on BLEU score.

\begin{table}[h]
\centering
\begin{tabular}{|l|l|l|l|l|l|}
\hline
Percentile & 20    & 40    & 60    & 80    & 99    \\ \hline
BLEU  & 0.09 & 0.20 & 0.34 & 0.53 & 0.82 \\ \hline
\end{tabular}
\caption{The distribution of BLEU score for conversations with more than 10 turns.}
\label{table:bleu_distribution}
\end{table}


\paragraph{Rating 1:} Conversations are given a rating of 1 if their BLEU score is less than or equal to 0.1.
We consider these conversations to be of low-quality.

\paragraph{Rating 2:} All the conversations that do not fit in rating 1 and 3 are marked with a rating of 2.

\paragraph{Rating 3:} Conversations are labeled with a rating of  3, only if the conversation has more than 12 turns and 
has a BLEU score larger than 0.587.
This threshold was calculated by summing the mean (0.385) and the standard deviation (0.202) of BLEU scores of the conversations that do not belong rating 1.

The average BLEU score for workers who have access to the document is 0.22 whereas the average BLEU score for the workers without access to the document is 0.03. 
This suggests that even if the workers had external knowledge about the movie, they have not extensively used it in the conversation. 
It also suggests that the workers with the document have not used the information from the document verbatim in the conversation.
Table \ref{table:dataset_statistics} shows the statistics on the total number of conversations, utterances, and average number of utterances per conversation and average length of utterances for all the three ratings.


\section{Models}
In this section we discuss models which can leverage the information from the document for generating responses. 
We explore generative models for this purpose.
Given a dataset $\boldsymbol{X} = \{x_0, \ldots, x_n\}$ of utterances in a conversation $C_i$, we consider two settings: (1) to generate a response $x_{i+1}$ when given only the current utterance $x_i$ and (2) to generate a response $x_{i+1}$ when given the corresponding section $s_i$ and the previous utterance $x_i$. 

\paragraph{Without section:}

We use the sequence-to-sequence model \cite{sutskever2014sequence} to build our baseline model. 
Formally, let $\boldsymbol{\theta}_{E}$ represent the parameters of the encoder. 
Then the representation $\boldsymbol{h}_{x_i}$ of the current utterance $x_i$ is given by:
\begin{equation}
\boldsymbol{h}_{x_i} = \text{Encoder}(\boldsymbol{x}_{i}; \boldsymbol{\theta}_{E})
\label{z_rep}
\end{equation}
Samples of $x_{i+1}$ are generated as follows:
\begin{equation}
p(\hat{\boldsymbol{x}} | \boldsymbol{h}_{x_i}) 
= \prod_{t} p(\hat{x}_{t}| \hat{\boldsymbol{x}}^{<t}, \boldsymbol{h}_{x_i})
\end{equation}
where, $\hat{\boldsymbol{x}}^{<t}$ are the tokens generated before $\hat{x}_t$. We also use global attention \cite{luong2015effective} with copy mechanism \cite{see2017get} to guide our generators to replace the unknown (UNK) tokens. We call this model \textbf{SEQ}.


\paragraph{With section:}

We extend the sequence-to-sequence framework to include the section $s_{i}$ corresponding the current turn. 
We use the same encoder to encode both the utterance and the section. 
We get the representation $\boldsymbol{h}_{x_i}$ of the current utterance $x_i$ using Eq. \ref{z_rep}.
The representation of the section is given by:
\begin{equation}
\boldsymbol{h}_{s_i} = \text{Encoder}(\boldsymbol{s}_i; \boldsymbol{\theta}_{E})
\end{equation}
The input at each time step $t$ to the generative model is given by $h_t = [x_{t-1};h_{s}]$, where $x_{t-1}$ is the embedding of the word at the previous time step.
We call this model \textbf{SEQS}.

\paragraph{Experimental Setup:}
For both SEQ and SEQS model, we use a two-layer bidirectional LSTM as the encoder and a LSTM as the decoder.
The dropout rate of the LSTM output is set to be 0.3. 
The size of hidden units for both LSTMs is 300. 
We set the word embedding size to be 100, since the size of vocabulary is relatively small\footnote{The total number of tokens is 46000, and we limit the vocabulary to be 10000 tokens.}. 
The models are trained with adam \cite{kingma2014adam} optimizer with learning rate 0.001 until they converge on the validation set for the perplexity criteria. 
We use beam search with size 5 for response generation.
We use all the data (i.e all the conversations regardless of the rating and scenario) for training and testing.
The proportion of train/validation/test split is 0.8/0.05/0.15.

\section{Evaluation}
In what follows, we first present an analysis of the dataset, then provide an automatic metric for evaluation of our models--perplexity and finally present the results of human evaluation of the generated responses for engagement and fluency.

\paragraph{Dataset analysis:}
\begin{table}[h!]
\centering
\begin{tabular}{|l|l|l|}
\hline
scenario & NW & LT \\ \hline
1        & 0.78   & 12.85        \\ \hline
2        & 5.84   & 117.12       \\ \hline                   
\end{tabular}
\caption{The results of data analysis. LT refers to the average length of $x_i$ in scenario 1 and $x_i, \ldots, x_{i+k}$ in scenario 2.}
\label{table:automated_eval1}
\end{table}

We perform two kinds of automated evaluation to investigate the usefulness of the document in the conversation. 
The first one is to investigate if the workers use the information from the document in the conversation.
The second analysis is to show that the document adds value to the conversation.
Let the set of tokens in the current utterance $x_i$ be $N$, the set of tokens in the current section $s_i$ be $M$, the set of tokens in 
the previous three utterances be $H$, and the set of stop words be $S$.
In scenario 1, we calculate the set operation (NW) as $| ((N \cap M) \setminus H) \setminus S|$.
Let the tokens that appear in all the utterances ($x_i, \ldots, x_{i+k}$) corresponding to the current section $s_i$ be $K$ and the tokens that appear in all the utterances ($x_i, \ldots, x_{i+p}$) corresponding to the previous section $s_{i-1}$ be $P$.
In scenario 2, we calculate the set operation (NW) as $| ((K \cap M) \setminus P) \setminus S|$.
The results in Table \ref{table:automated_eval1} show that people
use the information in the new sections and are not fixated on old sections. It also shows that they use the information to construct the responses. 

\paragraph{Perplexity:}
To automatically evaluate the fluency of the models, we use perplexity measure.
We build a language model on the train set of responses using ngrams up to an order of 3\footnote{We use the SRILM toolkit \cite{stolcke2002srilm}}.
The generated test responses achieve a perplexity of 21.8 for the SEQ model and \textbf{10.11} for the SEQS model. 
This indicates that including the sections of document helps in the generation process.


\subsection{Human Evaluation}
We also perform two kinds of human evaluations to evaluate the quality of predicted utterances -- engagement and fluency. These experiments are performed on Amazon Mechanical Turk. 

\paragraph{Engagement:}

We set up a pairwise comparison following \citet{bennett2005large} to evaluate the engagement of the generated responses. 
The test presents the chat history (1 utterance) and  then, in  random order, its corresponding response produced by the SEQ and SEQS models.
A third option ``No Preference'' was given to participants to mark no preference for either of the generated responses. 
The instruction given to the participants is ``Given the above chat history as context, you have to pick the one which can be best used as the response based on the engagingness.''
We randomly sample 90 responses from each model.
Each response was annotated by 3 unique workers and we take majority vote as the final label.
The result of the test is that SEQ generated responses were chosen only 36.4\% times as opposed to SEQS generated responses which were chosen \textbf{43.9\%} and the ``No Preference'' option was chosen 19.6\% of times.
This result shows the information from the sections improves the engagement of the generated responses.


\paragraph{Fluency:}
The workers were asked to evaluate the fluency of the generated response on a scale of 1 to 4, where 1 is unreadable and 4 is perfectly readable. 
We randomly select 120 generated responses from each model and each response was annotated by 3 unique workers.
The SEQ model got a low score of 2.88, contrast to the SEQS score of \textbf{3.84}.
This outcome demonstrates that the information in the section also helps in guiding the generator to produce fluent responses.


\section{Conclusion}
In this paper we introduce a crowd-sourced conversations dataset that is grounded in a predefined set of documents which is available for download \footnote{https://github.com/festvox/datasets-CMU\_DoG}.
We perform multiple automatic and human judgment based analysis to understand the value the information from the document provides to the generation of responses. 
The SEQS model which uses the information from the section to generate responses outperforms the SEQ model in the evaluation tasks of engagement, fluency and perplexity.



\section*{Acknowledgments}

This work was funded by a fellowship from Robert Bosch, and in part by Facebook Inc. and Microsoft  Corporation.
This work was performed as a part of The Conversational Intelligence Challenge (ConvAI, NIPS 2017) and we would like to thank the ConvAI team.
We  are  also  grateful  to  the  anonymous reviewers for their constructive feedback 
and to Carolyn Penstein Rose, Shivani Poddar, Sreecharan Sankaranarayanan, Samridhi Shree Choudhary and Zhou Yu for valuable discussions at earlier stages of this work.
\\

\bibliography{emnlp2018}
\bibliographystyle{acl_natbib_nourl}

\section{Appendix}
\label{sec:supplemental}

\subsection{Movie lists}
\begin{itemize}
\item Batman Begins
\item Bruce Almighty
\item Batman v Superman: Dawn of Justice
\item Catch me if you can
\item Despicable me (2010)
\item Dunkirk
\item Frozen (2013)
\item Home Alone
\item How to Train Your Dragon (2010)
\item The Imitation Game
\item Iron Man (2008)
\item Jaws
\item John Wick (2014)
\item La La Land
\item Maleficient
\item Mean Girls
\item Monsters University
\item Real Steel
\item The Avengers (2012)
\item The Blind Side
\item The Great Gatsby (2013)
\item The Inception
\item The Notebook
\item The Post
\item The Shape of Water
\item The Social Network
\item The Wolf of Wall Street
\item Toy Story
\item Wonder Woman
\item Zootopia
\end{itemize}

\subsection{Instructions given to the workers}
\label{sec:instructions_worker}

\subsubsection{Scenario 1: users with document}
\label{}
\begin{itemize}
\item The user you are pairing does not have the document you hold. Please read the document first.
\item Tell the other user what the movie is, and try to persuade the other user to watch/not to watch the movie using the information in the document. 
\item You should try to discuss the new paragraph when the document has changed. 
\item You will have 3 turns of conversation with your partner on each of the documents. 
\item You will be given 4 documents each containing a short paragraph. The new paragraph might show just beneath the previous document.
\item The next document will be loaded automatically after you finish 3 turns discussing the current document. 
\item You cannot use information you personally know that is not included there. You can use any information given in the document in the conversation.
\end{itemize}

\subsubsection{Scenario 1: users without document}
\label{}

\begin{itemize}
\item The other user will read a document about a movie.
\item If you are not told the name of the movie, try to ask the movie name.
\item After you are told the name of the movie, pretend you are interested in watching the movie, and try to gather all the information you need to make a decision whether to watch the movie in the end. 
\item You don’t have to tell the other user you decision in the end, but please share your mind at the feedback page.
\end{itemize}

\subsubsection{Scenario 2: both users with document}
\label{}

\begin{itemize}
\item The user you pair with has the same set of documents as yours. Please read the document first
\item Imagine you just watched this movie. Discuss the content in the document with the other user, and show whether you like/dislike the movie.
\item You should try to discuss the new paragraph when the document has changed.
\item You will have 3 turns of conversation with your partner on each of the documents.
\item You will be given 4 documents each containing a short paragraph. The new paragraph might show just beneath the previous document
\item The next document will be loaded automatically after you finish 3 turns discussing the current document.
\item You cannot use information you personally know that is not included there. You can use any information given in the document in the conversation.
\end{itemize}

\subsection{Post conversation survey questions}
\label{}

\subsubsection{For users with document}
Choose any:
\begin{itemize}
\item The document is understandable.
\item The other user is actively responding to me.
\item The conversation goes smoothly.
\end{itemize}
Choose one of the following:
\begin{itemize}
\item I have watched the movie before.
\item I have not watched the movie before.
\end{itemize}

\subsubsection{For users without document}
Choose any:
\begin{itemize}
\item The document is understandable.
\item The other user is actively responding to me.
\item The conversation goes smoothly.
\end{itemize}
Choose one of the following:
\begin{itemize}
\item I will watch the movie after the other user's introduction.
\item I will not watch the movie after the other user's introduction.
\end{itemize}

\subsection{Conversation Example 1}
\label{appendix:one_doc}

\begin{table*}[h!]
\centering
\begin{tabular}{ l l }
\hline
\multicolumn{2}{c}{Section 1} \\
\hline
\textbf{Name} & The inception \\
\textbf{Year} & 2009 \\
\textbf{Director} & Christopher Nolan \\
\textbf{Genre} & scientific \\
\textbf{Cast} & Leonardo DiCaprio as Dom Cobb, a professional thief who specializes in conning \\
& secrets from his victims by infiltrating their dreams. \\
& Joseph Gordon-Levitt as Arthur, Cobb's partner who manages and researches the missions. \\
& Ellen Page as Ariadne, a graduate student of architecture who is recruited to construct \\
& the various dreamscapes, which are described as mazes. \\
& Tom Hardy as Eames, a sharp-tongued associate of Cobb. \\
\textbf{Critical} & wildly ingenious chess game, the result is a knockout. \\
\textbf{Response} & DiCaprio, who has never been better as the tortured hero, draws you in with a love \\
& story that will appeal even to non-sci-fi fans. \\
& I found myself wishing Inception were weirder, further out the film is Nolan's \\
& labyrinth all the way, and it's gratifying to experience a summer movie with large \\
& visual ambitions and with nothing more or less on its mind than (as Shakespeare said) \\
& a dream that hath no bottom. \\
& Have no idea what so many people are raving about. It's as if someone went into their \\
& heads while they were sleeping and planted the idea that Inception is a visionary \\
& masterpiece and hold on Whoa! I think I get it. The movie is a metaphor for the power \\
& of delusional hype a metaphor for itself. \\
\textbf{Introd-} & Dominick Cobb and Arthur are extractors, who perform corporate espionage using an \\
\textbf{-uction} & experimental military technology to infiltrate the subconscious of their targets \\
& and extract valuable information through a shared dream world. Their latest target, \\
& Japanese businessman Saito, reveals that he arranged their mission himself to test \\
& Cobb for a seemingly impossible job: planting an idea in a person's subconscious, or \\
& inception. To break up the energy conglomerate of ailing competitor Maurice Fischer, \\
& Saito wants Cobb to convince Fischer's son and heir, Robert, to dissolve his father's\\
& company. \\
\textbf{Rating} & Rotten Tomatoes: 86\% and average: 8.1/10; IMDB: 8.8/10 \\
 \hline
\multicolumn{2}{c}{Conversation} \\
 \hline
user2: & Hey have you seen the inception? \\
user1: & No, I have not but have heard of it. What is it about \\
user2: & It's about  extractors that  perform experiments using military \\
& technology on people  to retrieve info about their targets.  \\
user1: & Sounds interesting do you know which actors are in  it? \\
user2: & I haven't watched it either or seen a preview.  Bu5 it's scifi \\
& so it might be good. Ugh Leonardo DiCaprio  is the main character  \\
user2: & He plays as Don Cobb \\
user1: & Oh okay, yeah I'm not  a big scifi fan but there are a few movies I still enjoy in that genre. \\
user1: & Is it a long movie? \\
user2: & Doesn't say how long it is.  \\
user2: & The Rotten Tomatoes score is  86\% \\
\end{tabular}
\caption{Utterances that corresponds to section 1 of the document in the example conversation 1.}
\label{table:c1s1}
\end{table*}

\begin{table*}[h!]
\centering
\begin{tabular}{ l l }
\hline
\multicolumn{2}{c}{Section 2} \\
\hline
\textbf{Scene 1} & When the elder Fischer dies in Sydney, Robert Fischer accompanies the \\
& body on a ten-hour flight back to Los Angeles, which the team (including \\
& Saito, who wants to verify their success) uses as an opportunity to sedate \\
& and take Fischer into a shared dream. At each dream level, the person \\
& generating the dream stays behind to set up a 'kick' that will be used to awaken \\
& the other sleeping team members from the deeper dream level; to be successful, \\
& these kicks must occur simultaneously at each dream level, a fact complicated due \\
& to the nature of time which flows much faster in each successive level. The first \\
& level is Yusuf's dream of a rainy Los Angeles. The team abducts Fischer, but they \\
& are attacked by armed projections from Fischer's subconscious, which has been \\
& specifically trained to defend him against such intruders. The team takes Fischer \\
& and a wounded Saito to a warehouse, where Cobb reveals that while dying in the \\
& dream would normally wake Saito up, the powerful sedatives needed to stabilize \\
& the multi-level dream will instead send a dying dreamer into 'limbo', a world of \\
& infinite subconscious from which escape is extremely difficult, if not almost \\
& impossible, and a dreamer risks forgetting they are in a dream. Despite these \\
& setbacks, the team continues with the mission. \\
 \hline
\multicolumn{2}{c}{Conversation} \\
 \hline
user1: & Wow, that's impressive. I like to look at Rotten Tomatoes when debating \\
& whether or not to see a movie. Do you know the director? \\
user2: & Something about Dom Cobb infiltrates peoples dreams in a dream world.  \\
user2: & The director is Christopher nolan \\
user2: & Heard of him?  \\
user2: & Wow I thought this was recent but it came out in 2009. \\
user1: & He directed The Dark Knight which I enjoy. Yeah, I know it's been out \\
& awhile but 2009 does seem to be a while back now. Time flies.  \\
user1: & Do you know if it won any awards? \\
user1: & or how much it made at the box office? \\
user2: & Oh wow I loved the dark night movies. And it doesn't say if it's won \\
& awards or how much at box office.  \\
user2: & A critic did say it could be "weirder" \\
\end{tabular}
\caption{Utterances that corresponds to section 2 of the document in the example conversation 1.}
\label{table:c1s2}
\end{table*}

\begin{table*}[h!]
\centering
\begin{tabular}{ l l }
\hline
\multicolumn{2}{c}{Section 3} \\
\hline
\textbf{Scene 2} & Cobb reveals to Ariadne that he and Mal went to Limbo while experimenting \\
& with the dream-sharing technology. Sedated for a few hours of real time, they \\
& spent fifty years in a dream constructing a world from their shared memories. \\
& When Mal refused to return to reality, Cobb used a rudimentary form of inception \\
& by reactivating her totem (an object dreamers use to distinguish dreams from reality) \\
& and reminding her subconscious that their world was not real. However, when she\\
& woke up, Mal still believed that she was dreaming. In an attempt to 'wake up' for \\
& real, Mal committed suicide and framed Cobb for her death to force him to do the \\
& same. Facing a murder charge, Cobb fled the U.S., leaving his children in the care \\
& of Professor Miles. \\
\hline
\multicolumn{2}{c}{Conversation} \\
\hline

user1: & The concept seems interesting and it has a good lead actor as well as director \\
& and reviews. I think it must be good. The plot does seem weird, that's for sure. \\
user2: & Tom Hardy is in the movie as the character Earnes. And yeah the plot is a bit strange.  \\
user2: & I might watch this movie now.  \\
user1: & I think I may as well. I can't say I've heard of Tom Hardy however. Is there \\
& any other supporting actors? \\
user2: & Oh Earnes is a sharp tongue associate of Cobb.  \\
user2: & Ellen Page \\
user1: & Oh, cool. I am familiar with her. She's in a number of good movies and is great. \\
user2: & She plays Ariadne , she is a graduate student that constructs the dreamscapes, \\
& they're like mazes.  \\
\end{tabular}
\caption{Utterances that corresponds to section 3 of the document in the example conversation 1.}
\label{table:c1s3}
\end{table*}

\begin{table*}[h!]
\centering
\begin{tabular}{ l l }
\hline
\multicolumn{2}{c}{Section 4} \\
\hline
\textbf{Scene 3} & Through his confession, Cobb makes peace with his guilt over Mal's death. \\
& Ariadne kills Mal's projection and wakes Fischer up with a kick. Revived \\
& at the mountain hospital, Fischer enters a safe room to discover and accept \\
& the planted idea: a projection of his dying father telling him to be his own \\
& man. While Cobb remains in Limbo to search for Saito, the other team members \\
& ride the synchronized kicks back to reality. Cobb eventually finds an aged Saito \\
& in Limbo and reminds him of their agreement. The dreamers all awake on the plane \\
& and Saito makes a phone call. Upon arrival at Los Angeles Airport, Cobb passes \\
& the U.S. immigration checkpoint and Professor Miles accompanies him to his home. \\
& Using his totem a spinning top that spins indefinitely in a dream world but falls \\
& over in reality Cobb conducts a test to prove that he is indeed in the real world,\\
& but he ignores its result and instead joins his children in the garden. \\
\hline
\multicolumn{2}{c}{Conversation} \\
\hline
user1: & Hmm interesting. Do you know if it's an action movie or mostly just scifi?  \\
user2: & Says scientific \\
user1: & Certainly seems unique. Do you know if it is based off a book or \\
& a previous work? \\
user2: & Something about at the end  he has trouble determining which is reality \\
& and which is a dream. It doesn't say it's based off anything.  \\
user1: & Sounds like it might be suspense/thriller as well as scifi which is cool. \\
& It seems pretty confusing but enticing. Makes me want to see it to try and \\
& figure it all out. \\
user2: & Yeah its like its got a bit of mystery too.  Trying to figure out \\
& what's real and what's not.  \\
user1: & I can't think of any other movie or even book that has a related story \\
& either which makes it very interesting. A very original concept. \\
user2: & Yeah well have great day. :) \\
\end{tabular}
\caption{Utterances that corresponds to section 4 of the document in the example conversation 1.}
\label{table:c1s4}
\end{table*}

\subsection{Conversation Example 2}
\label{appendix:two_docs}

\begin{table*}[h!]
\centering
\begin{tabular}{ l l }
\hline
\multicolumn{2}{c}{Section 1} \\
\hline
\textbf{Name} & The Shape of Water \\
\textbf{Year} & 2017\\
\textbf{Director} & Guillermo del Toro \\
\textbf{Genre} & Fantasy, Drama \\
\textbf{Cast} & Sally Hawkins as Elisa Esposito, a mute cleaner who works at a secret \\
& government laboratory. \\ 
& Michael Shannon as Colonel Richard Strickland, a corrupt military official, \\
& Richard Jenkins as Giles, Elisa's closeted neighbor and close friend who is a \\
& struggling advertising illustrator. \\
& Octavia Spencer as Zelda Delilah Fuller, Elisa's co-worker and friend who serves as \\ 
& her interpreter., \\
& Michael Stuhlbarg as Dimitri Mosenkov, a Soviet spy working as a scientist studying \\
& the creature, under the alias Dr. Robert Hoffstetler. \\
\textbf{Critical Response} & one of del Toro's most stunningly successful works, also a powerful vision \\
& of a creative master feeling totally, joyously free. \\
& Even as the film plunges into torment and tragedy, the core relationship between these \\
& two unlikely lovers holds us in thrall. Del Toro is a world-class film artist. \\
& There's no sense trying to analyze how he does it. \\
& The Shape of Water has tenderness uncommon to del Toro films. \\
& While The Shape of Water isn't groundbreaking, it is elegant and mesmerizing. \\
& refer Sally Hawkins' mute character as 'mentally handicapped' and for erroneously \\
& crediting actor Benicio del Toro as director. \\
\textbf{Introduction} & The Shape of Water is a 2017 American fantasy drama film directed by \\
& Guillermo del Toro and written by del Toro and Vanessa Taylor. It stars Sally Hawkins, \\
& Michael Shannon, Richard Jenkins, Doug Jones, Michael Stuhlbarg, and Octavia \\
& Spencer. Set in Baltimore in 1962, the story follows a mute custodian at a high-security \\
& government laboratory who falls in love with a captured humanoid amphibian creature. \\
\textbf{Rating} & Rotten Tomatoes: 92\% and average: 8.4/10 \\
      & Metacritic Score: 87/100\\
      & CinemaScore: A\\
 \hline
\multicolumn{2}{c}{Conversation} \\
 \hline
user1: & Hi \\ 
user2: & Hi \\ 
user2: & I thought The Shape of Water was one of Del Toro's best works. What about you? \\ 
user1: & Did you like the movie? \\ 
user1: & Yes, his style really extended the story. \\ 
user2: & I agree. He has a way with fantasy elements that really helped this story be truly \\
& beautiful. \\ 
user2: & It has a very high rating on rotten tomatoes, too. I don't always expect that with movies \\ 
& in this genre. \\
user1: & ally Hawkins acting was phenomenally expressive.  Didn't feel her character was \\
& mentally handicapped.\\
user2: & The characterization of her as such was definitely off the mark. 
\end{tabular}
\caption{Utterances that corresponds to section 1 of the document in the example conversation 2.}
\label{table:c2s1}
\end{table*}

\begin{table*}[h!]
\centering
\begin{tabular}{ l l }
\hline
\multicolumn{2}{c}{Section 2} \\
\hline
\textbf{Scene 1} & Elisa Esposito, who as an orphaned child, was found in a river with wounds on her neck, \\
& is mute, and communicates through sign language. She lives alone in an apartment above \\
& a cinema, and works as a cleaning-woman at a secret government laboratory in Baltimore \\
& at the height of the Cold War. Her friends are her closeted next-door neighbor Giles, a \\
& struggling advertising illustrator who shares a strong bond with her, and her co-worker \\
& Zelda, a woman who also serves as her interpreter at work. The facility receives a \\
& mysterious creature captured from a South American river by Colonel Richard Strickland, \\
& who is in charge of the project to study it. Curious about the creature, Elisa discovers \\
& it is a humanoid amphibian. She begins visiting him in secret, and the two form a close bond. \\
 \hline
\multicolumn{2}{c}{Conversation} \\
 \hline
user1: & Might as well label Giles too.   \\ 
user2: & haha. because he is closeted?  \\ 
user2: & Whoever made that comment was certainly not well informed and not politically \\
 & correct by any stretch.\\ 
user1: & I think Octavia Spencer should look for more roles set in the early 60s. \\
user2: & Do you think that the creature they find in the movie is supposed to be somehow\\
 & connected to the cold war?  \\ 
\end{tabular}
\caption{Utterances that corresponds to section 2 of the document in the example conversation 2.}
\label{table:c2s2}
\end{table*}

\begin{table*}[h!]
\centering
\begin{tabular}{ l l }
\hline
\multicolumn{2}{c}{Section 3} \\
\hline
\textbf{Scene 2} & Elisa keeps the creature in her bathtub, adding salt to the water to keep him alive. \\
& She plans to release the creature into a nearby canal when it will be opened to the ocean \\
& in several days' time. As part of his efforts to recover the creature, Strickland interrogates \\
& Elisa and Zelda, but the failure of his advances toward Elisa hampers his judgment, and he \\
&dismisses them. Back at the apartment, Giles discovers the creature devouring one of his \\
& cats, Pandora. Startled, the creature slashes Giles's arm and rushes out of the apartment. \\
& The creature gets as far as the cinema downstairs before Elisa finds him and returns him to \\
& her apartment. The creature touches Giles on his balding head and his wounded arm; the \\
& next morning, Giles discovers his hair has begun growing back and the wounds on his arm \\
& have healed. Elisa and the creature soon become romantically involved, having sex in her \\
& bathroom, which she at one point fills completely with water. \\
\hline
\multicolumn{2}{c}{Conversation} \\
\hline

user1: & Actually Del Toro does an incredible job showing working people. \\ 
user2: & That's an excellent point. \\ 
user1: & Yes, the Cold War invented the Russians, I kind of thought it also represented technology \\ 
 & in general. \\  
user2: & That makes perfect sense. \\ 
user2: & I really like that Eliza chose to keep the creature in her bathtub.  \\
user1: & It was interesting that neither power treated the monster well. \\
user1: & Yes the magical realism was truly magical ... easy to suspend disbelief.
\end{tabular}
\caption{Utterances that corresponds to section 3 of the document in the example conversation 2.}
\label{table:c2s3}
\end{table*}

\begin{table*}[h!]
\centering
\begin{tabular}{ l l }
\hline
\multicolumn{2}{c}{Section 4} \\
\hline
\textbf{Scene 3} & Hoyt gives Strickland an ultimatum, asking him to recover the creature within \\
& 36 hours. Meanwhile, Mosenkov is told by his handlers that he will be extracted in two \\
& days. As the planned release date approaches, the creature's health starts deteriorating. \\
& Mosenkov leaves to rendezvous with his handlers, with Strickland tailing him. At the \\
& rendezvous, Mosenkov is shot by one of his handlers, but Strickland shoots the handlers \\
& dead and then tortures Mosenkov for information. Mosenkov implicates Elisa and Zelda \\
& before dying from his wounds. Strickland then threatens Zelda in her home, causing her \\
& terrified husband to reveal that Elisa had been keeping the creature. Strickland searches \\
& Elisa's apartment and finds a calendar note revealing when and where she plans to release \\
& him. At the canal, Elisa and Giles bid farewell to the creature, but Strickland arrives and \\
& attacks them all. Strickland knocks Giles down and shoots the creature and Elisa, who \\
& both appear to die. However, the creature heals himself and slashes Strickland's throat, \\
& killing him. As police arrive on the scene with Zelda, the creature takes Elisa and \\
& jumps into the canal, where, deep under water, he heals her. When he applies his healing \\
& touch to the scars on her neck, she starts to breathe through gills. In a closing voiceover \\
& narration, Giles conveys his belief that Elisa lived 'happily ever after' with the creature. \\
\hline
\multicolumn{2}{c}{Conversation} \\
\hline
user2: & Yes. I think it was beautiful that the creature essentially had healing power. \\
user1: & Del Toro does well with violence. \\
user1: & The ending was suspenseful, without being over the top. \\
user2: & What a powerful ending. Even though it was obviously a pure fantasy scenario, there was \\
& so much real emotion. \\
user2: & He does do well with violence. I've noticed that in all of his movies. \\
user2: & Del Toro is one of my favorite directors. \\
user1: & Yes, happy endings usually feel fake.  This one felt great. \\
user2: & Totally. It felt like what should have happened, rather than just a sappy pretend ending that \\
& was forced on the viewer. \\
user1: & Mine too.  Evidently Hollywood is starting to agree. \\
user2: & It took a while, but yes, finally. \\
user1: & It really appeared to be filmed in Baltimore.  Installation looked so authentic. \\
user2: & Do you know where it was actually filmed?  \\
user1: & No. Can you imagine soaking in that pool? \\
user2: & :) \\
user1: & Would make a great tourist draw.  \\
user2: & That would be amazing! What a great idea! \\
user2: & Haven't we completed the amount of discussion needed yet? \\
user1: & Place looked like a cross between a nuclear power plant and an aquarium. \\
& I think we hit all the points mentioned. \\
\end{tabular}
\caption{Utterances that corresponds to section 4 of the document in the example conversation 2.}
\label{table:c2s4}
\end{table*}

\end{document}